\pdfoutput=1
\documentclass[11pt]{article}

\usepackage[]{acl}

\usepackage{times}
\usepackage{latexsym}
\usepackage{graphicx}
\usepackage{caption}
\usepackage{siunitx}
\usepackage{enumitem}
\usepackage{amsfonts}
\usepackage{adjustbox}
\usepackage[T1]{fontenc}

\usepackage[utf8]{inputenc}

\usepackage{microtype}

\title{Voxel-informed Language Grounding}
\author{Rodolfo Corona \qquad Shizhan Zhu \qquad Dan Klein \qquad Trevor Darrell \\
  Computer Science Division, University of California, Berkeley \\
  \texttt{\{rcorona, shizhan\_zhu, klein, trevordarrell\}@berkeley.edu}}

\begin{document}
\maketitle

\begin{abstract}
Natural language applied to natural 2D images describes a fundamentally 3D world. 
We present the Voxel-informed Language Grounder (VLG), a language grounding model that leverages \textit{3D geometric information} in the form of voxel maps derived from the visual input using a volumetric reconstruction model.  
We show that VLG significantly improves grounding accuracy on SNARE~\cite{thomason2021language}, an object reference game task.
At the time of writing, VLG holds the top place on the SNARE leaderboard,\footnote{\href{https://github.com/snaredataset/snare\#leaderboard}{https://github.com/snaredataset/snare\#leaderboard}} achieving SOTA results with a 2.0\% absolute improvement.  
\end{abstract}

\section{Introduction}

Embodied robotic agents hold great potential for providing assistive technologies in home environments~\cite{pineau2003towards}, and
natural language provides an intuitive interface for users to interact with such systems~\cite{andreas2020task}. 
For these systems to be effective, they must be able to reliably ground language in perception~\cite{bisk2020experience,bender2020climbing}.

\begin{figure}
    \centering
    \begin{adjustbox}{max size={0.99\columnwidth}{0.99\textheight}}
    \includegraphics{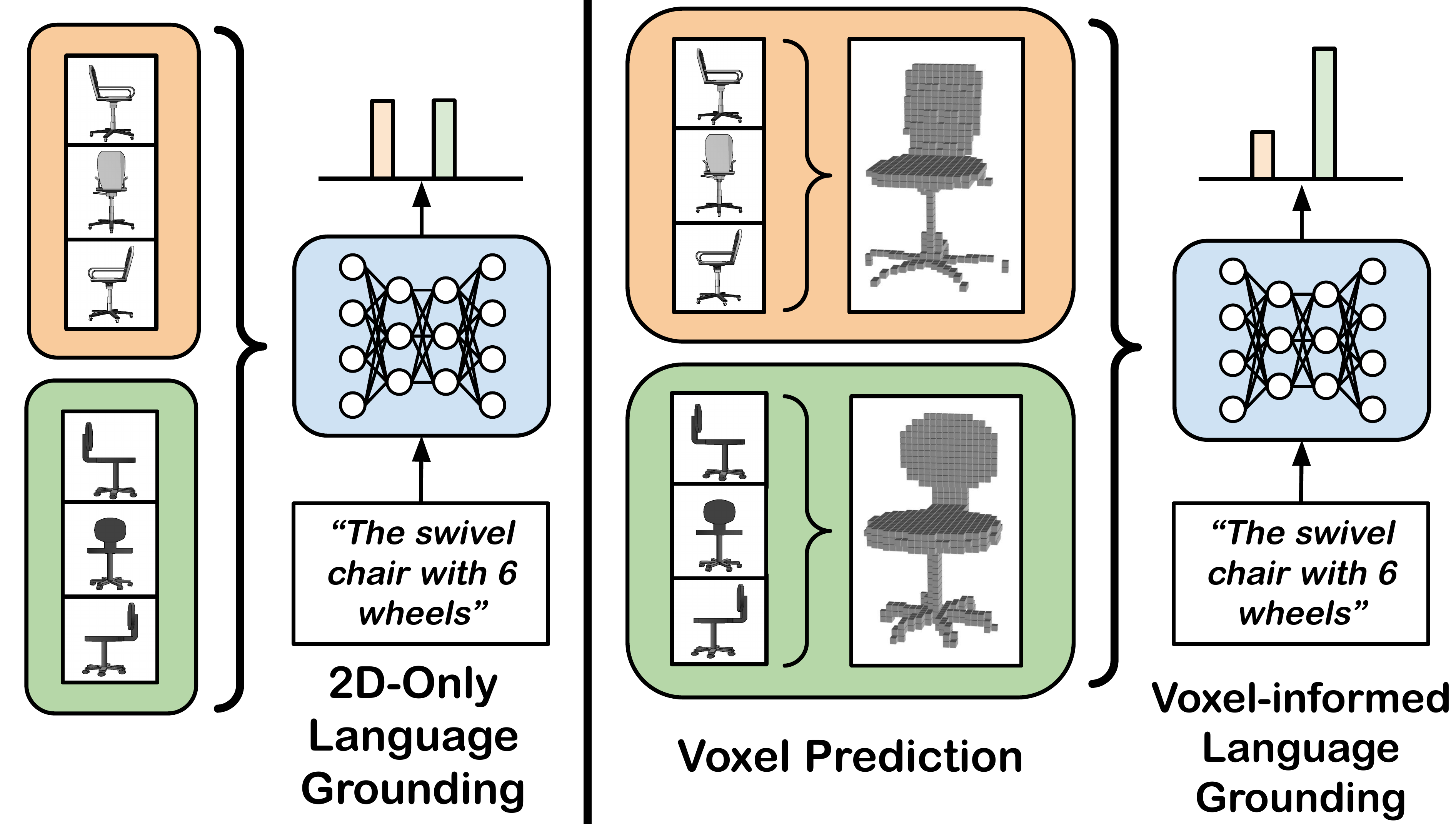}%
    \end{adjustbox}
    \captionsetup{width=.95\linewidth}
    \caption{\textbf{Voxel-informed Language Grounder.} Our VLG model leverages explicit 3D information by inferring volumetric voxel maps from input images, allowing the agent to reason jointly over the geometric and visual properties of objects when grounding.}
    \label{fig:teaser}
\end{figure}

Despite typically being paired with 2D images, natural language that is grounded in vision describes a fundamentally 3D world.
For example, consider the grounding task in Figure \ref{fig:teaser}, where the agent must select a target chair against a distractor given the description ``the swivel chair with 6 wheels.'' 
Although the agent is provided with multiple images revealing all of the wheels on each chair, it must be able to properly aggregate information across images to successfully differentiate them, something that requires reasoning about their \textit{3D geometry} at some level.

In this work, we show how language grounding performance may be improved by leveraging 3D prior knowledge. 
Our model, Voxel-informed Language Grounder (VLG), extracts 3D voxel maps using a pre-trained \textit{volumetric reconstruction model}, which it fuses with multimodal features from a large-scale vision and language model in order to reason jointly over the visual and 3D geometric properties of objects. 

We focus our investigation within the context of SNARE~\cite{thomason2021language}, an object reference game where an agent must ground natural language describing common household objects by their geometric and visual properties, showing that grounding accuracy significantly improves by incorporating information from predicted 3D volumes of objects.
At the time of writing, VLG achieves SOTA performance on SNARE, attaining an absolute improvement of 2.0\% over the next closest baseline. 
Code to replicate our results is publicly available.\footnote{\small\url{https://github.com/rcorona/voxel_informed_language_grounding}}

\section{Related Work}

Prior work has studied deriving structured representations from images to scaffold language grounding. 
However, a majority of systems use representations such as 2D regions of interest~\cite{anderson2018bottom,wang2020language} or symbolic graph-based representations~\cite{hudson2019learning,kulkarni2013babytalk}, which do not encode 3D properties of objects.

Most prior work tying language to 3D representations has largely focused on generating 3D structures conditioned on language, rather than using them as intermediate representations for language grounding as we do here. 
Specifically, prior work has performed language conditioned generation at the scene~\cite{chang2014learning,chang2015shapenet}, pose~\cite{ahuja2019language2pose,lin2018generating}, or object~\cite{chen2018text2shape} level.
More recently, a line of work has explored referring expression grounding in 3D by mapping referring expressions of objects to 3D bounding boxes localizing them in point clouds of indoor scenes~\cite{achlioptas2020referit_3d,chen2020scanrefer,Zhao_2021_ICCV,roh2022languagerefer}.
Standard approaches follow a two-tiered process where an object proposal system will first provide bounding boxes for candidate objects, and a scoring module will then compute a compatibility score between each box and the referring expression in order to ground it. 
At a more granular level, \citet{koo2021partglot} learn alignments from language to object parts by training agents on a reference game over point cloud representations of objects. 

In contrast, in this work we focus on augmenting language grounding over 2D RGB images using structured 3D representations derived from them. 
For the task of visual language navigation, prior work has shown how a persistent 3D semantic map may be used as an intermediate representation to aid in selecting navigational waypoints~\cite{chaplot2020object,blukis2021persistent}.
The semantic maps, however, represent entire scenes with individual voxels representing object categories, rather than their geometry.  
In this work, we show how a more granular occupancy map representing objects' geometry can improve language grounding performance.

Closest to our work is that of \citet{prabhudesai2020embodied}, which presents a method for mapping language to 3D features within scenes from the CLEVR~\cite{johnson2017clevr} dataset.
Their system generates 3D feature maps inferred from images and then grounds language directly to 3D bounding boxes or coordinates.
Their method assumes, however, that dependency parse trees are provided for the natural language inputs, and it is trained with supervised alignments between noun phrases and the 3D representations, which VLG does not require. 

\section{Voxel-informed Language Grounder}

\begin{figure*}
    \centering
    \begin{adjustbox}{max size={0.99\textwidth}{0.99\textheight}}
    \includegraphics{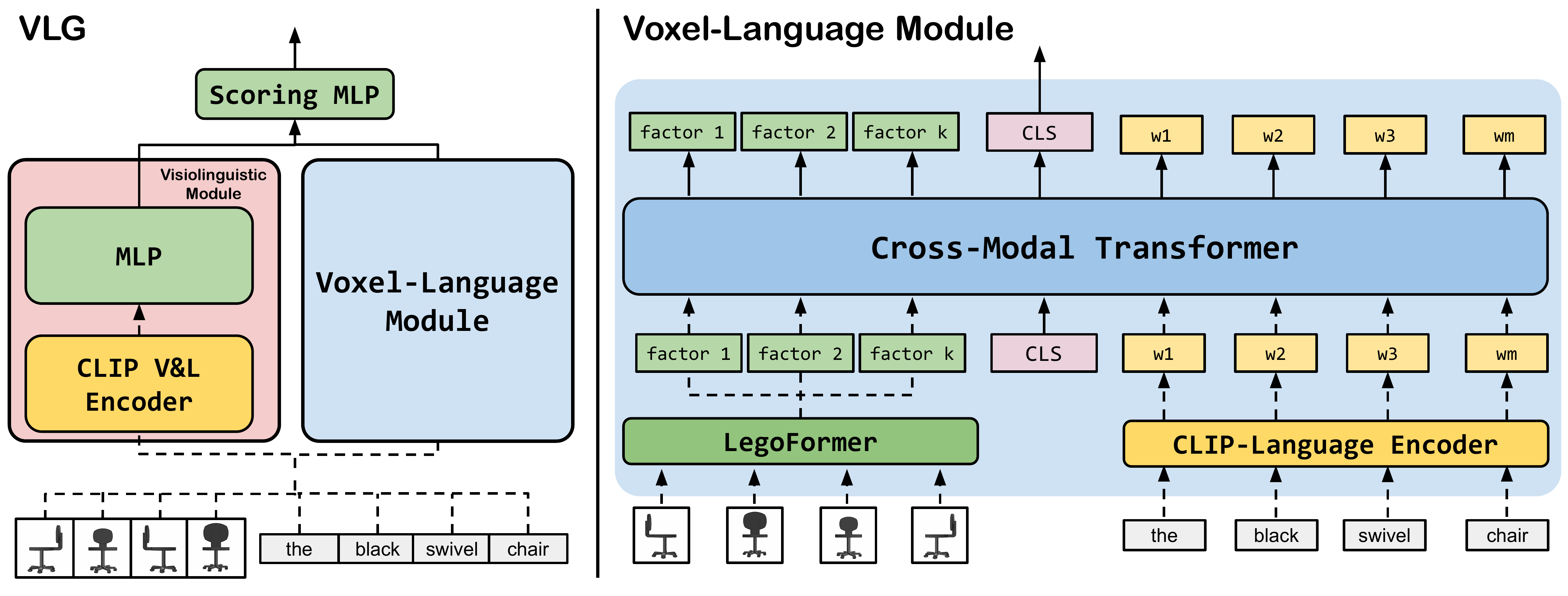}%
    \end{adjustbox}
    \captionsetup{width=.99\textwidth}
    \caption{\textbf{VLG Architecture.} (Left) Our VLG model consists of a visiolinguistic module which produces a joint embedding for text and images using CLIP~\cite{radford2021learning} and a voxel-language module for jointly embedding language and volumetric maps. (Right) The voxel-language module uses a cross modal transformer to fuse word embeddings from CLIP with voxel map factors extracted from LegoFormer~\cite{yagubbayli2021legoformer}. During training, gradients only flow through solid lines.}
    \label{fig:model}
\end{figure*}

We consider a task where an agent must correctly predict a target object $v^t$ against a distractor $v^c$ given a natural language description $w^t=\{w_1, ..., w_m\}$ of the target. 
For each object, the agent is provided with $n$ 2D views $v=\{x_1, ..., x_n\}$, $x_i\in \mathbb{R}^{3\times W\times H}$.

An agent for this task is represented by a scoring function $s(v, w)\in [0,1]$, computing the compatibility between the target description and the 2D views of each object, and is used to select the maximally scoring candidate.
We first use unimodal encoders to encode the language description into $e_w=h(w)$ and the object view images into a single aggregate visual embedding $e_v = g(v)$ before fusing them with a visiolinguistic module $e_{vw}=f_{vw}\left([e_v;e_w]\right)$.
Prior approaches to this problem~\cite{thomason2021language} directly input this fused representation to a scoring module to produce a score $s(e_{vw})$.
They do not explicitly reason about the 3D properties of the observed objects, requiring the models to learn them implicitly. 

In contrast, our Voxel-informed Language Grounder augments the scoring function $s$ with explicit 3D volumetric information $e_o=o(v)$ extracted from a pre-trained multiview reconstruction model.
The volumetric information (in the form of a factorization of a voxel occupancy map in $\mathbb{R}^{W\times H\times D}$) is first fused into a joint representation with the language using a multimodal voxel-language module $e_{ow}=f_{ow}([e_o;e_w])$. 
The scoring function then produces a score based on all three modalities $s([e_{vw};e_{ow}])$.

\subsection{Model Architecture}

VLG (Figure \ref{fig:model}) consists of two branches: a visiolinguistic module for fusing language and 2D RGB features, and a voxel-language module for fusing language with 3D volumetric features. A scoring function is then used to reason jointly over the output of the two branches, producing a compatibility score. 
\\
\\
\textbf{Visiolinguistic Module.} The architecture of our visiolinguistic module $f_{vw}$ (left panel, Figure \ref{fig:model}) largely mirrors the architecture of MATCH from \citet{thomason2021language}.
A pre-trained CLIP-ViT~\cite{radford2021learning} model is used to encode the language description and view images into vectors in $\mathbb{R}^{512}$. 
The image embeddings are max-pooled and concatenated to the description embedding before being passed into an MLP which generates a fused representation. 
\\
\\
\textbf{Voxel-Language Module.} We use representations extracted from a ShapeNet~\cite{shapenet2015,wu20153d} pre-trained LegoFormerM~\cite{yagubbayli2021legoformer}, a multi-view 3D volumetric reconstruction model, as input to our voxel-language module $f_{ow}$.
LegoFormer is a transformer~\cite{vaswani2017attention} based model whose decoder generates volumetric maps factorized into 12 parts. 
Each object factor is represented by a set of three vectors $x,y,z\in \mathbb{R}^{32}$, which we concatenate to use as input tokens for our voxel-language module. 
A triple cross-product over $x,y,z$ may be used to recover a 3D volume $\mathcal{V}\in  \mathbb{R}^{32 \times 32 \times 32}$ for each factor. 
The full volume for the object is generated by aggregating the factor volumes through a sum operation. 
For more details on LegoFormer, we refer the reader to \citet{yagubbayli2021legoformer}.
We use a cross-modal transformer~\cite{vaswani2017attention} encoder to fuse the language and object factors (Figure \ref{fig:model}, right). 
The cross-modal transformer takes as input language tokens, in the form of CLIP word embeddings, and the 12 object factors output by the LegoFormer decoder, which contain the inferred geometric occupancy information of the object. 
We use a CLS token as an aggregate representation of the language and object factors.
\\
\\
\textbf{Scoring Function.} The scoring function is represented by an MLP which takes as input the concatenation of the visiolinguistic module output and the cross-modal transformer's CLS token.

\section{Language Grounding Evaluation}\label{sec:eval}

\begin{table*}[ht!]
    \centering
    \begin{tabular}{c}
    \phantom{000000}\textbf{VALIDATION}\phantom{00000000000000000000000}\textbf{TEST}\\
    \end{tabular}
    \begin{tabular}{l|ccc|ccc}
    Model & Visual\phantom{(0.0)} & Blind\phantom{(0.0)} & All\phantom{(0.0)} & Visual\phantom{(0.0)} & Blind\phantom{(0.0)} & All\phantom{(0.0)} \\
    \hline
    ViLBERT & 89.5\phantom{(0.0))} & 76.6\phantom{(0.0))} & 83.1\phantom{(0.0))} & 80.2\phantom{(0.0))} & \textbf{73.0}\phantom{(0.0))} & 76.6\phantom{(0.0))} \\
    MATCH & 89.2 (0.9)  & 75.2 (0.7) & 82.2 (0.4) & 83.9 (0.5) & 68.7 (0.9) & 76.5 (0.5) \\
    MATCH$^*$ & 90.6 (0.4) & 75.7 (1.2) & 83.2 (0.8) & -\phantom{0000} & -\phantom{0000} & -\phantom{0000} \\
    LAGOR & 89.8 (0.4) & 75.3 (0.7)& 82.6 (0.4) & 84.3 (0.4) & 69.4 (0.5) & 77.0 (0.5) \\
    LAGOR$^*$ & 89.8 (0.5) & 75.0 (0.4) & 82.5 (0.1) & -\phantom{0000} & -\phantom{0000} & -\phantom{0000} \\
    VLG (Ours) & \textbf{91.2} (0.4) & \textbf{78.4}\rlap{$^\dagger$} (0.7) & \textbf{84.9}\rlap{$^\dagger$} (0.3) & \textbf{86.0}\phantom{(0.0)} & 71.7\phantom{(0.0)} & \textbf{79.0}\phantom{(0.0)} \\
    \hline
    \end{tabular}
    \caption{\textbf{SNARE Benchmark Performance.} Object reference game accuracy on the SNARE task across validation and test sets. Performance on models with an asterisk are our replications of the baselines in \citet{thomason2021language}. Standard deviations over 3 seeds are shown in parentheses. MATCH corresponds to the max-pool variant from \citet{thomason2021language} since no test set results are provided for the mean-pool variant. Our VLG model achieves the best overall performance. Due to leaderboard submission restrictions, we were not able to get test set results for the MATCH$^*$ and LAGOR$^*$ replications. $\dagger$ denotes statistical significance over replicated models with $p < 0.1$.}
    \label{tab:snare_results}
\end{table*}

\textbf{Evaluation.} We test our method on the SNARE benchmark~\cite{thomason2021language}.
SNARE is a language grounding dataset which augments ACRONYM~\cite{eppner2021acronym}, a grasping dataset built off of ShapeNetSem~\cite{savva2015semgeo,chang2015shapenet}, with natural language annotations of objects.

SNARE presents an object reference game where an agent must correctly guess a target object against a distractor. 
In each instance of the game, the agent is provided with a language description of the target as well as multiple 2D views of each object.
SNARE differentiates between \textbf{visual} and \textbf{blindfolded} object descriptions.
Visual descriptions primarily include attributes such as \textit{name}, \textit{shape}, and \textit{color} (e.g. ``classic armchair with white seat''). 
In contrast, blindfolded descriptions include attributes such as \textit{shape} and \textit{parts} (e.g. ``oval back and vertical legs''). 
The train/validation/test sets were generated by splitting over (207 / 7 / 48) ShapeNetSem object categories, respectively containing (6,153 / 371 / 1,357) unique object instances and (39,104 / 2,304 / 8,751) object pairings with referring expressions. 
Renderings are provided for each object instance over 8 canonical viewing angles. 

Because ShapeNet and ShapeNetSem represent different splits of the broader ShapeNet database, we pre-train the LegoFormerM model on a modified dataset to avoid dataset leakage. 
Specifically, any objects which appear in both datasets are re-assigned within the pre-training dataset used to train LegoFormerM to match its split assignment from SNARE. 

ShapeNetSem images are resized to $224\times224$ when inputting them to LegoFormerM in order to match its ShapeNet pre-training conditions. 
\\
\\
\textbf{Baselines.} We compare VLG against the set of models provided with SNARE.\footnote{\small\url{https://github.com/snaredataset/snare}}
All SNARE baselines except \textbf{ViLBERT} use a CLIP ViT-B/32~\cite{radford2021learning} backbone for encoding both images and language descriptions:

\begin{itemize}
    \item[] \textbf{MATCH} first uses CLIP-ViT to embed the language description as well as each of the 8 view images. Next, the view embeddings are mean-pooled and concatenated to the description embedding. Finally, a learned MLP is used over the concatenated feature vector in order to produce a final compatibility score.
    
    \item[] \textbf{ViLBERT} fine-tunes a 12-in-1~\cite{lu202012} pre-trained ViLBERT\cite{lu2019vilbert} as the backbone for MATCH instead of using CLIP-ViT. Each object is presented to ViLBERT in the form of a single tiled image containing all 14 views from ShapeNetSem, instead of just the canonical 8 presented in the standard task. ViLBERT tokenizes images by extracting features from image regions, with the ground truth bounding boxes for each region (i.e. view) being provided. Because this baseline is not open-source, we report the original numbers from ~\citet{thomason2021language}.  
    
    \item[] \textbf{LAGOR} (\textbf{La}nguage \textbf{G}rounding through \textbf{O}bject \textbf{R}otation) fine-tunes a pre-trained MATCH module and is additionally regularized through the auxiliary task of predicting the canonical viewing angle of individual view images, which it predicts using an added output MLP head. Following ~\citet{thomason2021language}, the LAGOR baseline is only provided with 2 random views of each object both during training and inference. 
\end{itemize}

For more details on the baseline models, we refer the reader to \citet{thomason2021language}.
\\
\\
\textbf{Training Details}. Apart from the dataset split re-assignments mentioned in Section \ref{sec:eval}, we use the code\footnote{\small\url{https://github.com/faridyagubbayli/LegoFormer}} and hyperparameters presented by~\citet{yagubbayli2021legoformer} to train LegoFormerM. 

For training on SNARE, we follow ~\citet{thomason2021language} and train all models with a smoothed binary cross-entropy loss~\cite{achlioptas2019shapeglot}.

We train each model for 75 epochs, reporting performance of the best performing checkpoint on the validation set. 
For our replication of the SNARE MATCH and LAGOR baselines, we use the code and hyperparameters provided by \citet{thomason2021language}. 
For all variants of our VLG model we use the AdamW~\cite{loshchilov2017decoupled} optimizer with a learning rate of 1e-3 and a linear learning rate warmup of 10K steps.

\section{Results}\label{sec:results}

We present test set performance for VLG and the SNARE baselines reported by~\citet{thomason2021language}. 
We also present average performance for trained models over 3 seeds with standard deviations on the validation set. 

\subsection{Comparison to SOTA}

In Table \ref{tab:snare_results} we can observe reference game performance for all models. 
VLG achieves SOTA performance with an absolute improvement on the test set of 2.0\% over LAGOR, the next best leaderboard model. 
Although there is a general improvement of 1.7\% in \textbf{visual} reference grounding, there is an improvement of 2.3\% in \textbf{blindfolded} (denoted as \textbf{Blind} in tables to conserve space) reference grounding. 
This suggests that the injected 3D information provides a greater boost for disambiguating between examples referring to geometric properties of target objects. 
VLG generally improves over all baselines and conditions for blindfolded examples, with the exception of ViLBERT, which may be due to the additional information ViLBERT receives in the form of 14 viewing angles of each object instead of 8. 
Improvements on the Blind and All conditions of the validation set are statistically significant over replicated models with $p < 0.1$ under a Welch's two-tailed $t$-test.

\subsection{Ablation Study}

\begin{table}[]
\tabcolsep=0.11cm
    \begin{tabular}{c|ccc}
        Model & Visual & Blind & All \\
        \hline
        VGG16 & \textbf{91.4} (0.5) & 76.5 (0.9) & 84.0 (0.2) \\
        MLP  & 91.1 (0.8) & 77.9 (0.9) & 84.6 (0.1) \\
        no-CLIP  & 71.0 (0.6) & 65.8 (0.7) & 68.4 (0.1) \\
        VLG & 91.2 (0.4) & \textbf{78.4} (0.7) & \textbf{84.9} (0.3) \\
    \end{tabular}
    \caption{\textbf{Ablation Study.} SNARE reference game accuracy across ablations of our model on the validation set. We show performance when replacing LegoformerM object factors with \textbf{VGG16} features, replacing the cross-modal transformer with an \textbf{MLP}, and when foregoing the use of CLIP features (\textbf{no-CLIP}).}
    \label{tab:ablations}
\end{table}

We present a variety of ablations on the validation set to investigate the contributions of each piece of our model. 
All results can be observed in Table \ref{tab:ablations}.
\\
\\
\textbf{VGG16 Embeddings.} LegoFormer uses an ImageNet~\cite{deng2009imagenet} pre-trained VGG16~\cite{simonyan2014very} as a backbone for extracting visual representations, which is a different dataset and pre-training task than what the CLIP-ViT image encoder is trained on.
This presents a confounding factor which we ablate by performing an experiment feeding our model's scoring function VGG16 features directly instead of LegoFormer object factors (VGG16 in Table \ref{tab:ablations}). 
Despite getting comparable results to VGG16 on visual reference grounding, VLG provides a clear improvement in blindfolded (and therefore overall) reference performance, suggesting that the extracted 3D information is useful for grounding more geometrically based language descriptions, with the VGG16 features being largely redundant in terms of visual signal. 
\\
\\
\textbf{Architecture.} We ablate the contribution of our cross-modal transformer branch by comparing it against an MLP mirroring the structure of the SNARE MATCH baseline.
This model (MLP in Table \ref{tab:ablations}) max-pools the LegoFormer object factors and concatenates the result to the CLIP visual and language features before passing them to an MLP scoring function. 
The MLP model overall outperforms the SNARE baselines from Table \ref{tab:snare_results}, highlighting the usefulness of the 3D information for grounding, but does not result in as large an improvement as the cross-modal transformer. 
This suggests that the transformer is better able at integrating information from the multi-view input. 
\\
\\
\textbf{CLIP Visual Embeddings.} Finally, we evaluate the contribution of the visiolinguistic branch of the model by removing it and only using the cross-modal transformer over language and object factors. 
As may be observed, there is a large drop in performance (16.5\% overall), particularly for visual references (20.2\%). 
These results suggest that maintaining visual information such as color and texture is critical for performing well on this task, since the LegoFormer outputs contain only volumetric occupancy information.  

\section{Discussion}

We have presented the Voxel-informed Language Grounder (VLG), a model which leverages explicit 3D information from predicted volumetric voxel maps to improve language grounding performance. 
VLG achieves SOTA results on SNARE, and ablations demonstrate the effectiveness of using this 3D information for grounding. 
We hope this paper may inspire future work on integrating structured 3D representations into language grounding tasks. 

\section*{Acknowledgements}

We would like to thank Karttikeya Mangalam and Nikita Kitaev for their helpful advice and discussions on transformer models. Mohit Shridhar and Jesse Thomason for their help with setting up SNARE. And thanks to the anonymous reviewers for their constructive feedback. 
This work was supported by DARPA under the SemaFor program (HR00112020054). The content does not necessarily reflect the position or the policy of the government, and no official endorsement should be inferred.
RC is supported by an NSF Graduate Research Fellowship. 

\bibliography{acl_latex}
\bibliographystyle{acl_natbib}

\end{document}